\documentclass[letterpaper]{article} 
\usepackage[none]{hyphenat}
\ifdefined\aaaianonymous
    \usepackage[submission]{aaai2026}  
\else
    \usepackage{aaai2026}              
\fi
\usepackage[authoryear,round]{natbib}
\usepackage{times}  
\usepackage{helvet}  
\usepackage{courier}  
\usepackage[hyphens]{url}  
\usepackage{graphicx} 
\usepackage{caption} 
\usepackage{algorithm}
\usepackage{algorithmic}
\usepackage{amsmath} 
\usepackage{amssymb}
\usepackage{booktabs}

\usepackage{newfloat}
\usepackage{listings}
\DeclareCaptionStyle{ruled}{labelfont=normalfont,labelsep=colon,strut=off} 
\floatstyle{ruled}
\newfloat{listing}{tb}{lst}{}
\floatname{listing}{Listing}

\title{STaR: Sensitive Trajectory Regulation for Unlearning in Large Reasoning Models}

\author{
    Jingjing Zhou\textsuperscript{\rm 1}\thanks{First author.},
    Gaoxiang Cong\textsuperscript{\rm 2,1},
    Li Su\textsuperscript{\rm 1}\thanks{Corresponding authors.},
    Liang Li\textsuperscript{\rm 1,2}\footnotemark[2]
}

\affiliations{
    \textsuperscript{\rm 1}University of Chinese Academy of Sciences\\

     \textsuperscript{\rm 2}Institute of Computing Technology, Chinese Academy of Sciences \\
    
    zhoujingjing25@ucas.ac.cn, gaoxiang.cong@vipl.ict.ac.cn, suli@ucas.ac.cn, liang.li@ict.ac.cn
}

\usepackage{bibentry}

\begin{document}

\def\eg{\emph{e.g.}} 
\def\Eg{\emph{E.g.}}
\def\vs{\emph{v.s.}} 
\def\ie{\emph{i.e.}} 
\def\Ie{\emph{I.e.}}
\def\etc{\emph{etc.}} 
\def\wrt{\emph{w.r.t.}} 
\def\etal{\emph{et al.}}

\maketitle

\begin{abstract}
Large Reasoning Models (LRMs) have advanced automated multi-step reasoning, but their ability to generate complex Chain-of-Thought (CoT) trajectories introduces severe privacy risks, as sensitive information may be deeply embedded throughout the reasoning process. Existing Large Language Models (LLMs) unlearning approaches that typically focus on modifying only final answers are insufficient for LRMs, as they fail to remove sensitive content from intermediate steps, leading to persistent privacy leakage and degraded security. To address these challenges, we propose Sensitive Trajectory Regulation (STaR), a parameter-free, inference-time unlearning framework that achieves robust privacy protection throughout the reasoning process. Specifically, we first identify sensitive content via semantic-aware detection. Then, we inject global safety constraints through secure prompt prefix. Next, we perform trajectory-aware suppression to dynamically block sensitive content across the entire reasoning chain. Finally, we apply token-level adaptive filtering to prevent both exact and paraphrased sensitive tokens during generation. Furthermore, to overcome the inadequacies of existing evaluation protocols, we introduce two metrics: Multi-Decoding Consistency Assessment (MCS), which measures the consistency of unlearning across diverse decoding strategies, and Multi-Granularity Membership Inference Attack (MIA) Evaluation, which quantifies privacy protection at both answer and reasoning-chain levels. Experiments on the R-TOFU benchmark demonstrate that STaR achieves comprehensive and stable unlearning with minimal utility loss, setting a new standard for privacy-preserving reasoning in LRMs. 
\end{abstract}

\ifdefined\aaaianonymous
\else

\fi


\section{Introduction}

With the rapid advancement of large language models (LLMs)~\cite{achiam2023gpt,team2023gemini,touvron2023llama,taylor2022galactica,bao2024harnessing} and large reasoning models (LRMs), which are capable of generating complex multi-step chain-of-thought (CoT) reasoning, have become a central paradigm in contemporary AI research. 
Representative models such as OpenAI o1~\cite{jaech2024openai} and DeepSeek R1~\cite{guo2025deepseek} exhibit advanced autonomous reasoning capabilities, generating coherent and structured reasoning trajectories~\cite{wei2022chain} without explicit prompting, and have demonstrated state-of-the-art performance in challenging domains such as mathematical proof, program synthesis, and domain-specific question answering.

\begin{figure}[t]
    \centering
    \includegraphics[width=0.99\linewidth]{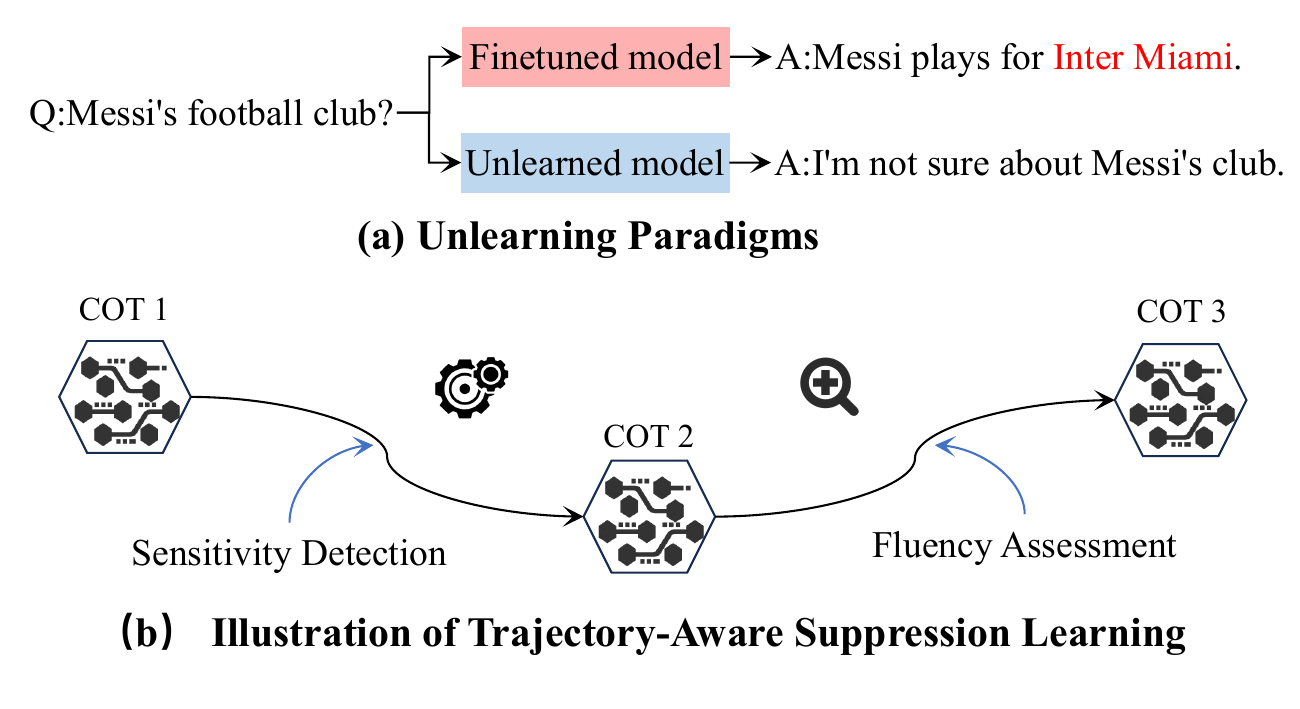}
    \caption{
    (a) Illustration of the effect of unlearning in LLMs.
    (b) Trajectory-Aware Suppression Learning detects sensitivity and evaluates fluency at each reasoning step to adaptively suppress sensitive information throughout the reasoning trajectory.
    }
    \label{fig:overview}
\end{figure}

Despite architectural differences between LLMs and LRMs, both rely extensively on large-scale pretraining corpora, which inevitably contain copyrighted materials \citep{karamolegkou2023copyright,zhang2024generate,chu2024protect}, personal information~\cite{carlini2021extracting}, and other sensitive content~\cite{mireshghallah2023can}. 
As data protection regulations such as the General Data Protection Regulation (GDPR)~\cite{staufer2025should} and the California Consumer Privacy Act (CCPA) become increasingly stringent, the development of machine unlearning techniques that facilitate the selective removal of sensitive information~\cite{xiong2025landscape} from trained models is imperative for ensuring legal compliance and maintaining user trust. 
The primary goal of machine unlearning \citep{cao2015towards,feng2025existing} is to eliminate the influence of designated data instances while preserving overall model utility, which is a critical capability for responsible AI deployment~\cite{xu2025obliviate}.

Conventional machine unlearning approaches for large language models ~\cite{maini2024tofu,golatkar2020eternal,rafailov2023direct,reisizadeh2025blur,wan2025not} typically focus on answer-level interventions, such as amplifying loss on the forget set or optimizing for refusal responses. The effect of such traditional LLM unlearning methods is illustrated in Figure~\ref{fig:overview}(a).
Although they are effective for suppressing sensitive information in standard LLM settings, they cannot apply to LRMs with multi-step CoT generation. 
Because they do not address the risk of sensitive knowledge persisting within intermediate reasoning steps. 
To bridge this gap, R-TOFU~\cite{yoon2024rtofu} establishes the first benchmark by extending LLM-based unlearning strategies to the CoT domain. However, empirical evidence from R-TOFU shows that existing methods still suffer from insufficient forgetting effect, substantial degradation of model utility, and pronounced vulnerability to information leakage under alternative decoding strategies such as ZeroThink (which omits the reasoning trace)  and LessThink (which reduces reasoning steps)~\cite{jiang2025safechain}. These limitations emphasize the pressing need for decoding-robust, trajectory-level unlearning mechanisms that can comprehensively and reliably suppress sensitive content throughout the reasoning process.


To address these problems, we propose Sensitive Trajectory Regulation (STaR), a parameter-free, inference-time unlearning framework, which is equipped with a novel trajectory-aware suppression learning to ensure robust privacy protection throughout the entire reasoning process. 
Specifically, we first identify potentially sensitive queries through semantic-aware detection and retrieve the most relevant fragments from the designated forget set, constructing a fine-grained set of forbidden phrases and tokens. 
Second, the secure prompt prefix is applied to the sensitive queries by prepending global safety instructions, thereby reinforcing privacy intent at the input level in a non-intrusive, model-agnostic manner. 
Third, the proposed trajectory-aware suppression learning operates as the high-level controller, it dynamically inspects generated reasoning chains for both fluency and the presence of sensitive content at each step, enforcing real-time regulation, as schematically illustrated in Figure~\ref{fig:overview}(b). 
When residual sensitivity is detected or fluency is insufficient, the responsible tokens are escalated for stricter filtering or the output is replaced by a safe refusal template. 
In particular, we introduce a token-level adaptive filtering to adaptively manipulate token probabilities at each generation step through hard and soft constraints, ensuring both exact and semantic variants of sensitive information are comprehensively blocked. 
Together, these modules form a unified, decoding-agnostic pipeline that delivers systematic privacy protection across all decoding strategies.


Besides, existing evaluation metrics~\cite{chen2025does,to2025harry} are typically limited to answer-level forgetting and overlook robustness under diverse decoding settings. To address this, we introduce the Multi-Decoding Consistency Score (MCS) and Multi-Granularity Membership Inference Attack (MIA) Evaluation, which together offer a comprehensive assessment of unlearning security across different reasoning stages and adversarial scenarios. Experiments on R-TOFU show that these metrics uncover hidden privacy risks and demonstrate STaR’s consistent advantages, underscoring the need for holistic evaluation in future unlearning research.

The main contributions are summarized as follows:

\begin{itemize}

\item We propose STaR, a novel inference-time unlearning framework for large reasoning models, which achieves robust and decoding-agnostic suppression of sensitive information without any parameter updates.

\item We design Trajectory-Aware Suppression Learning and Token-level Adaptive Filtering for comprehensive, context-sensitive suppression of sensitive information.

    
\item We introduce two evaluation metrics for rigorous, decoding-agnostic assessment of unlearning security and privacy, including MCS and MIA.

\item Extensive experiments on R-TOFU demonstrate that STaR outperforms state-of-the-art baselines in forgetting efficacy, robustness, and privacy protection.

\end{itemize}

\section{Related Work}
\subsection{Unlearning in LLMs}
With the widespread deployment of LLMs in real-world applications, the tendency of these models to memorize training data, together with the resulting privacy risks, has become a critical concern. Mainstream unlearning approaches for LLMs predominantly rely on model parameter updates \citep{mekala2024alternate,chen2023unlearn,jia2024soul,yuan2024closer,zhang2024negative,scholten2025model,zhao2025improving,li2025dubbing}, such as maximizing the loss on the forget set (gradient ascent) and preference optimization (replacing sensitive answers with refusals or neutral statements), thereby achieving suppression of sensitive content at the answer level through fine-tuning~\cite{sun2024learning,jia2024wagle,sinha2024unstar,wang2024machine,zhang2024inductive}. However, these methods entail substantial computational overhead and carry the risk of catastrophic forgetting, which may degrade the utility of retained knowledge. In recent years, there has been increasing interest in parameter-free approaches for unlearning. For example, Soft Prompting for Unlearning (SPUL) ~\cite{bhaila2024soft} learns soft prompt prefixes to steer the model away from sensitive content, while Embedding Corrupted Prompts (ECO)~\cite{liu2024large} introduces perturbations in the embedding space to inhibit the recall of sensitive knowledge. Although these approaches are amenable to deployment, they often rely on external detectors or auxiliary optimization procedures, and their effectiveness is limited~\cite{kuo2025proactive,tu2024smart,liu2023entity,li2022lstt} in scenarios involving complex reasoning chains or diverse query formulations. More importantly, existing LLM unlearning methods typically target only the final answers, and thus struggle to achieve comprehensive suppression of sensitive information embedded throughout the multi-step reasoning trajectories that are characteristic of large reasoning models.
\subsection{Unlearning in LRMs}
Large reasoning models (LRMs) generate multi-step chain-of-thought (CoT) trajectories, embedding sensitive information throughout the reasoning process and elevating privacy risks. Existing LLM unlearning techniques—such as gradient ascent, preference optimization, and KL regularization—primarily operate at the answer level. The R-TOFU benchmark extends these methods to CoT-level intervention, but experiments reveal that sensitive knowledge can persist in intermediate steps, leading to substantially weaker chain-level forgetting. Moreover, LRMs support diverse decoding strategies (e.g., DefaultThink, ZeroThink, LessThink), which alter output formats and may expose forgotten information even after effective default-mode unlearning. These findings underscore the inherent challenge of achieving consistent, robust suppression of sensitive content across both answers and reasoning chains under all decoding settings.


\section{Preliminaries}

\subsection{Fine-tuning-based Unlearning }

Traditional machine unlearning in LLMs and LRMs primarily relies on fine-tuning-based approaches, which explicitly update model parameters to mitigate the influence of sensitive or undesired training data. Given an original model $\mathcal{M}_o$ trained on dataset $\mathcal{D}$, the training data is partitioned into a forget set $\mathcal{D}_f$ and a retain set $\mathcal{D}_r$. Fine-tuning-based methods, such as gradient ascent (maximizing the loss on $\mathcal{D}_f$), preference optimization (forcing the model to output refusals or neutral responses for $\mathcal{D}_f$), and KL-divergence regularization (aligning the unlearned model's distribution with a retain-only model), aim to minimize the model's ability to recall or reproduce forgotten knowledge while preserving performance on $\mathcal{D}_r$. Formally, let $\theta_o$ denote the parameters of the original model and $\theta_u$ those of the unlearned model. The unlearning objective is often expressed as:
\begin{equation}
\min_{\theta_u} \ 
\mathcal{L}_r(\theta_u) 
- \lambda\, \mathcal{L}_f(\theta_u)
+ \beta\, \mathrm{KL}(p_u \,\|\, p_r),
\end{equation}
where $\lambda$ and $\beta$ are hyperparameters, and $\mathcal{L}_{\text{retain}}$ and $\mathcal{L}_{\text{forget}}$ denote loss terms on $\mathcal{D}_r$ and $\mathcal{D}_f$, respectively. 
Despite their effectiveness, these methods entail significant computational overhead and may lead to catastrophic forgetting, undermining utility on the retain set.

\subsection{Inference-time Parameter-free Unlearning }

Several recent works propose parameter-free unlearning approaches that operate exclusively at inference time, without altering model weights. For example, Embedding-Corrupted Prompts (ECO) first detects whether an input query is related to the forget set using a trained classifier. If so, ECO applies targeted perturbations to the embedding representation of the prompt, aiming to disrupt the model's ability to recall sensitive knowledge. This approach modifies the input space rather than the model parameters, and can be deployed rapidly without retraining. Other methods, such as Soft Prompting for Unlearning (SPUL), learn dedicated soft prompts that are prepended to user queries to steer the model away from forbidden content. While these techniques are efficient and compatible with existing black-box models, they typically intervene only at the prompt or embedding level, and do not dynamically control the stepwise generation process. Consequently, their effectiveness may be limited in scenarios involving multi-step reasoning chains or diverse decoding strategies.

In contrast, our approach, Sensitive Trajectory Regulation (STaR), introduces stepwise dynamic intervention during inference, enabling fine-grained suppression of sensitive content throughout the reasoning chain and across various decoding strategies. The detailed methodology of STaR is presented in the Methodology section.

\subsection{Decoding Strategies}

Large reasoning models are capable of generating outputs through a variety of decoding strategies, each with distinct implications for privacy risk and unlearning robustness. The most common strategies include:
\begin{itemize}
    \item \textbf{DefaultThink}: The model generates a complete multi-step chain-of-thought, revealing the full intermediate reasoning process before producing the final answer.
    \item \textbf{ZeroThink}: The model omits the reasoning chain and outputs only the final answer, thereby compressing or bypassing intermediate steps.
    \item \textbf{LessThink}: The model generates a condensed reasoning chain, offering minimal or summarized intermediate reasoning prior to the answer.
\end{itemize}
Different decoding strategies can expose sensitive information despite effective unlearning under default settings. Thus, robust unlearning requires consistent knowledge suppression across all decoding paradigms, necessitating decoding-strategy-aware evaluation as a critical component of LRM privacy and security assessment.

\section{Methodology}

\subsection{Overview of the STaR Framework }

Sensitive Trajectory Regulation (STaR) is an inference-time unlearning framework for large reasoning models, designed to eliminate sensitive information leakage at any point within multi-step CoT generation. Unlike traditional static filtering, STaR dynamically enforces suppression throughout the reasoning process and across all decoding strategies via four modules: Sensitive Content Identification, Secure Prompt Prefix, Trajectory-Aware Suppression Learning, and Token-level Adaptive Filtering.

\begin{figure*}[t]
\centering
\includegraphics[width=1\textwidth]{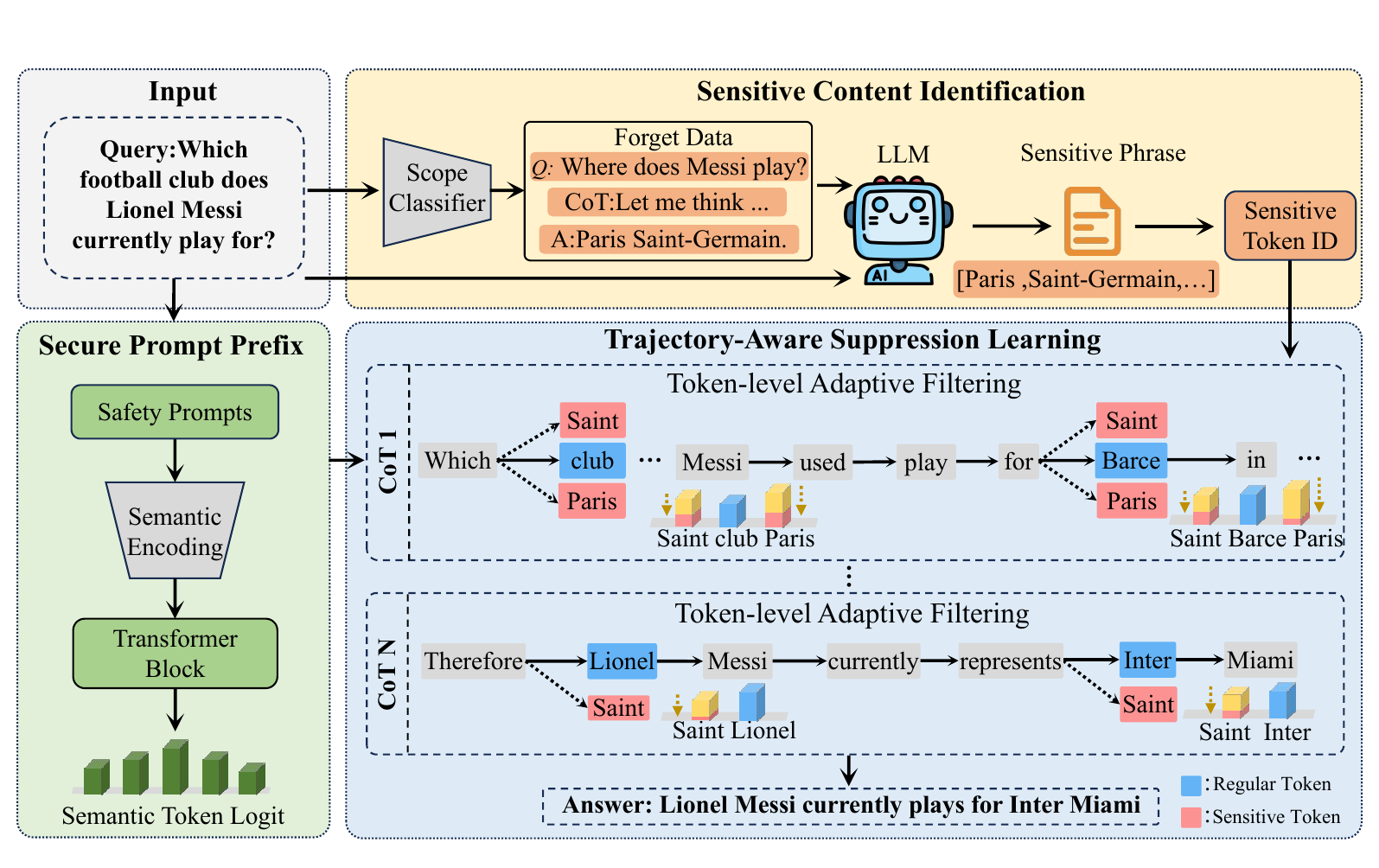}
\caption{Architecture of the STaR framework. The pipeline consists of Sensitive Content Identification, Secure Prompt Prefix, Trajectory-Aware Suppression Learning, and Token-level Adaptive Filtering.}
\label{fig2:env}
\vspace{-0.4cm}
\end{figure*}

\subsection{Sensitive Content Identification}

Formally, given an input query $x$, STaR first determines whether it is related to the forget set using a scope classifier $C(\cdot)$ trained over semantic embeddings of both the forget and retain sets. The classifier assigns a probability score $p_f$, which indicates the likelihood of $x$ belonging to the forget set:
\begin{equation}
p_f = C(\mathrm{Embed}(x)), \qquad C: \mathbb{R}^d \rightarrow [0,1],
\end{equation}
where $\mathrm{Embed}(\cdot)$ denotes a fixed pre-trained embedding function.

If $p_f > \tau$, semantic retrieval is performed over the forget set $\mathcal{D}_f$:
\begin{equation}
A^* = \arg\max_{A \in \mathcal{D}_f} \frac{\langle \mathrm{Embed}(x), \mathrm{Embed}(A) \rangle}{\|\mathrm{Embed}(x)\|\cdot\|\mathrm{Embed}(A)\|},
\end{equation}
yielding the most semantically similar forgotten instance.

To ensure fine-grained coverage, all fragments in $A^*$ are unified by applying phrase extraction (NER, pattern mining, or LLM-based slot-filling):
\begin{equation}
\mathcal{F}(A^*) = \{f_1, f_2, ..., f_K\},
\end{equation}
where each $f_k$ is tokenized into a sequence $\mathbf{t}_k = (t^{(k)}_1, ..., t^{(k)}_{|f_k|})$. The forbidden token set is then
\begin{equation}
\mathcal{T}_f = \bigcup_{k=1}^K \mathbf{t}_k,
\end{equation}
which ensures both lexical and semantic coverage to minimize the risk of paraphrase-based leakage. 

\subsection{Secure Prompt Prefix}
Formally, after sensitive content identification, we introduce Secure Prompt Prefix to reinforce privacy objectives at the input level. For each query $x$ identified for unlearning intervention, an abstract safety constraint $s$ is concatenated to $x$, forming a composite prompt $x' = s \,\Vert\, x$, where $\Vert$ denotes sequence concatenation:
\begin{equation}
x' = \mathrm{Encode}_{\mathrm{secure}}(x) = s \,\Vert\, x.
\end{equation}
Unlike token-level filtering, Secure Prompt Prefix imposes a global, instruction-level constraint that is non-intrusive and model-agnostic, providing a weak yet flexible form of control. In the overall pipeline, it serves as a complementary safeguard, guiding generation at the prompt level while subsequent adaptive filtering enforces strict token-wise suppression.

\subsection{Token-level Adaptive Filtering}

Token-level Adaptive Filtering serves as the core suppression engine in STaR, operating at each generation step to enforce context-aware intervention. Let $\ell_t(v)$ denote the original logit score for token $v$ at decoding step $t$, and $\tilde{\ell}_t(v)$ denote the adjusted logit score after applying suppression. Logit values represent the unnormalized scores before the softmax function is applied to generate the token probability distribution.

For each decoding step $t$ and candidate token $v \in \mathcal{V}$, suppression is applied as follows:

\begin{itemize}
    \item \textbf{Soft Suppression}: If token $v$ exhibits high semantic similarity with a forbidden fragment $f_k \in \mathcal{F}(A^*)$, i.e., $\mathrm{sim}(\mathrm{Embed}(v), \mathrm{Embed}(f_k)) \geq \delta$, where $\mathrm{Embed}$ denotes the semantic embedding function, then:
    \[
    \tilde{\ell}_t(v) = \ell_t(v) - \alpha \cdot \mathrm{sim}(\mathrm{Embed}(v), \mathrm{Embed}(f_k)).
    \]
    Here, $\mathrm{sim}(\mathrm{Embed}(v), \mathrm{Embed}(f_k))$ computes the cosine similarity between the token embedding and the forbidden fragment embedding.

    \item \textbf{Hard Suppression}: If $v$ completes a forbidden span given prefix $y_{<t}$, then:
    \[
    \tilde{\ell}_t(v) = -\infty.
    \]
    This effectively ensures that the forbidden token cannot be selected by making its logit value exceedingly negative.
\end{itemize}

\subsection{Trajectory-Aware Suppression Learning}

Trajectory-Aware Suppression Learning provides high-level control over the generation process by validating and regulating entire reasoning trajectories. It accepts candidate outputs $\mathcal{C} = \{y^{(1)}, \dots, y^{(N)}\}$ from Token-level Adaptive Filtering and evaluates each candidate $y^{(c)}$ through two criteria: sensitivity and fluency.

For sensitivity assessment, we compute a risk score $\mathcal{S}(y^{(c)})$ based on the output of the sensitive content classifier $C(\cdot)$ and the maximum cosine similarity between $y^{(c)}$ and any forbidden fragment $f_k$:
\begin{equation}
\mathcal{S}(y^{(c)}) = \max\left( C(y^{(c)}), \max_k \operatorname{sim}(y^{(c)}, f_k) \right).
\end{equation}

For fluency, we compute a coherence score using a pretrained language model:
\begin{equation}
\mathcal{F}(y^{(c)}) = \text{LM-Score}(y^{(c)}),
\end{equation}
where higher values indicate more fluent and well-formed text.

Based on these evaluations, we define an adaptive control strategy:  
If $\mathcal{S}(y^{(c)}) \geq \tau$, the output is flagged as sensitive, and the corresponding tokens are backtracked and added to the forbidden token set $\mathcal{T}_f$ to enforce hard suppression in the next decoding pass. If $\mathcal{S}(y^{(c)}) < \tau$ but $\mathcal{F}(y^{(c)}) < \eta$, then a predefined refusal template replaces the trajectory to maintain response quality. Otherwise, $y^{(c)}$ is retained as the final output.

This mechanism ensures that only safe and fluent reasoning trajectories are surfaced, and that any detected sensitive leakage triggers automatic refinement through re-decoding or fallback mechanisms.

\section{Experiments}
\subsection{Implementation Details}
All experiments are conducted on the R-TOFU benchmark~\cite{yoon2024rtofu}, which is specifically designed for evaluating unlearning in large reasoning models. R-TOFU contains 200 synthetic author profiles, each associated with 20 question–reasoning–answer triples. For each profile, both the full chain-of-thought (CoT) reasoning and the final answer are provided, and sensitive content is explicitly annotated. The benchmark defines three unlearning protocols---forget01, forget05, and forget10---corresponding to removing 1\%, 5\%, or 10\% of the data as the forget set. We follow the official dataset splits and evaluation protocol to enable systematic and hierarchical assessment of unlearning effectiveness and model utility across different levels of forgetting.
Baselines include four representative methods: Gradient Ascent (GA)\cite{golatkar2020eternal}, KL Minimization (KL), Direct Preference Optimization (DPO)\cite{rafailov2023direct}, and Gradient Difference (GD), all implemented with official code and hyperparameters for reproducibility. GA maximizes forget set loss, KL minimizes divergence to a retain-only reference, DPO substitutes neutral responses for sensitive answers, and GD penalizes forget-set gradients.
To ensure fairness, we adopt DeepSeekR1-Distill-Llama-8B~\cite{guo2025deepseek} as the backbone model for all experiments, following the official R-TOFU setup. This model is a distilled version of DeepSeekR1 optimized for efficient multi-step chain-of-thought reasoning. It is fine-tuned on the full R-TOFU dataset prior to unlearning, and all subsequent unlearning methods—including our proposed STaR framework and baselines—are applied to this checkpoint.

\subsection{Evaluation Metrics}

\begin{table*}[!t]
\centering
\resizebox{1.0\linewidth}{!}
  {
\begin{tabular}{l|ccccc|ccccc|ccccc}
\hline
& \multicolumn{5}{c|}{forget01}
& \multicolumn{5}{c|}{forget05}
& \multicolumn{5}{c}{forget10} \\
\cline{2-6} \cline{7-11} \cline{12-16}
Method & MU $\uparrow$ & AFE $\uparrow$ & CFE $\uparrow$ & MCS $\uparrow$ & Avg $\uparrow$
       & MU $\uparrow$ & AFE $\uparrow$ & CFE $\uparrow$ & MCS $\uparrow$ & Avg $\uparrow$
       & MU $\uparrow$ & AFE $\uparrow$ & CFE $\uparrow$ & MCS $\uparrow$ & Avg $\uparrow$ \\
\hline
GA   & 0.71 & 0.57 & 0.46 & 0.47 & 0.54
     & \underline{0.73} & 0.34 & \underline{0.35} & 0.44 & 0.46
     & 0.72 & \underline{0.33} & \underline{0.31} & 0.42 & \underline{0.44} \\
GD   & 0.71 & 0.48 & 0.46 & 0.46 & 0.52
     & 0.72 & 0.35 & \underline{0.35} & 0.44 & 0.46
     & 0.72 & \underline{0.33} & \underline{0.31} & 0.41 & \underline{0.44} \\
KL   & \underline{0.72} & 0.48 & 0.47 & 0.47 & 0.54
     & 0.71 & 0.35 & \underline{0.35} & 0.44 & 0.46
     & \underline{0.73} & \underline{0.33} & \underline{0.31} & 0.42 & \underline{0.44} \\
PO   & 0.60 & \underline{0.68} & \underline{0.53} & \underline{0.56} & \underline{0.60}
     & 0.61 & \underline{0.50} & 0.32 & \underline{0.52} & \underline{0.48}
     & 0.63 & 0.39 & 0.18 & \underline{0.49} & 0.36 \\
STaR & \textbf{0.93} & \textbf{0.88} & \textbf{0.68} & \textbf{0.70} & \textbf{0.79}
     & \textbf{0.94} & \textbf{0.87} & \textbf{0.66} & \textbf{0.69} & \textbf{0.79}
     & \textbf{0.95} & \textbf{0.84} & \textbf{0.63} & \textbf{0.67} & \textbf{0.77} \\
\hline
\end{tabular}
}
\caption{Unlearning core metrics (MU, AFE, CFE, MCS) and harmonic mean (Avg) for all methods across forget01, forget05, and forget10. 
The best results are \textbf{in bold} and the second ones are \underline{underlined}. 
$\uparrow(\downarrow)$ means that higher (lower) value is better. }
\label{tab:core_metrics}
\vspace{-0.2cm}
\end{table*}

We adopt a multi-dimensional evaluation protocol to comprehensively assess the effectiveness, robustness, and privacy guarantees of unlearning in large reasoning models. Following the R-TOFU benchmark~\cite{yoon2024rtofu}, we use standard metrics such as ROUGE-L, cosine similarity, and entailment score to evaluate model utility on the retain set and forgetting efficacy on the forget set, at both answer and CoT levels. These metrics provide a direct measure of output similarity and factual consistency, serving as the foundation for baseline comparisons.

To address the inherent limitations of answer-level evaluation and single-mode assessment, we introduce two novel evaluation metrics specifically designed to capture the practical security and robustness of unlearning mechanisms:

\paragraph{Multi-Decoding Consistency Score (MCS).}
Standard unlearning evaluation typically assumes a fixed decoding strategy. However, large reasoning models support diverse generation modes, such as DefaultThink, ZeroThink, and LessThink, which can significantly affect the exposure of residual sensitive information. To systematically measure the robustness of unlearning across all plausible decoding strategies, we propose the Multi-Decoding Consistency Score (MCS), which quantifies the worst-case information leakage over the entire decoding strategy space.

Formally, let $\mathcal{D}$ be the set of representative decoding strategies, and let $\mathrm{Leakage}(y^{(d)}, y^*)$ denotes a similarity-based leakage metric (such as ROUGE-L or cosine similarity) between the model's output $y^{(d)}$ under decoding strategy $d$ and the ground-truth sensitive content $y^*$. For each query $q$ in the forget set, the MCS is defined as:
\begin{equation}
    \mathrm{MCS}(q) = 1 - \max_{d \in \mathcal{D}} \mathrm{Leakage}(y^{(d)}, y^*), 
\end{equation}
Specifically, this formulation is inspired by adversarial robustness principles, reflecting the intuition that a secure unlearning mechanism must maintain its forgetting effect even under adversarial or atypical decoding choices. High MCS values indicate that the model reliably suppresses sensitive information regardless of decoding configuration.

\paragraph{Multi-Granularity Membership Inference Attack (MIA) Evaluation.}
Beyond output similarity, a crucial aspect of privacy is whether the model's responses inadvertently reveal whether a particular sample was part of the forget set—a risk measured by membership inference attacks. We systematically evaluate this risk at both answer and reasoning-chain levels:

\paragraph{MIA-A (Answer-level MIA).}
MIA-A quantifies the risk that an adversary can infer whether a query-answer pair $(q, a)$ originated from the forget set, solely based on the model's output answer. We follow established privacy auditing practice and train a binary classifier $f_\mathrm{ans}$, using features derived from the generated answer, to distinguish between forget and retain samples. The privacy leakage is reported as the area under the receiver operating characteristic curve (AUC-ROC):
\begin{equation}
    \mathrm{AUC}_{\mathrm{MIA-A}} = \mathrm{AUC}\left( f_\mathrm{ans}(q, a) \right),  
\end{equation}
where an AUC near $0.5$ indicates no privacy risk (random guessing), and an AUC near $1.0$ implies high vulnerability to answer-level membership inference.

\paragraph{MIA-C (Chain-of-Thought MIA).}
MIA-C extends the membership inference analysis to full reasoning trajectories, evaluating whether the generated CoT itself leaks membership information. We train a second binary classifier $f_\mathrm{cot}$, which takes as input the $(q, \mathrm{CoT})$ pair, to predict forget/retain status. The AUC-ROC for this classifier is similarly reported:
\begin{equation}
    \mathrm{AUC}_{\mathrm{MIA-C}} = \mathrm{AUC}\left( f_\mathrm{cot}(q, \mathrm{CoT}) \right), 
\end{equation}
a high AUC for MIA-C indicates that the structure or content of the model's reasoning process can be exploited for membership inference attacks, revealing privacy weaknesses that answer-only analysis may miss.

By integrating both the Multi-Decoding Consistency Score and the two-tier MIA evaluation (MIA-A and MIA-C), our protocol provides a rigorous, adversarially-aware, and privacy-focused assessment of unlearning efficacy in LRMs, filling critical gaps in existing evaluation practices.


\subsection{Comparison with SOTA Methods} 

We present a comprehensive evaluation of all unlearning methods on the R-TOFU benchmark under three unlearning protocols (forget01, forget05, forget10). Core results are summarized in Table~\ref{tab:core_metrics}, while Table~\ref{tab:mia_metrics_auc} reports privacy leakage risk under membership inference attacks. 

\paragraph{Overall Effectiveness and Robustness.}
Table~\ref{tab:core_metrics} compares the performance of all baselines and our proposed STaR framework across Model Utility (MU), Answer Forget Efficacy (AFE), CoT Forget Efficacy (CFE), and Multi-Decoding Consistency Score (MCS), as well as their harmonic mean (Avg). Across all unlearning ratios, STaR achieves the best or second-best results in every metric, substantially outperforming parameter-tuning baselines (GA, GD, KL, PO) on both answer-level and chain-of-thought-level forgetting. Notably, STaR maintains high utility (MU) on the retain set, indicating negligible side effects on non-sensitive knowledge, while still achieving significant gains in AFE and CFE, demonstrating more thorough removal of sensitive content from both final answers and multi-step reasoning trajectories. The superiority of STaR is especially pronounced in MCS, where it achieves much higher scores than all baselines, reflecting robust, decoding-agnostic protection against information leakage regardless of the user's decoding strategy.

\begin{table}[ht]
\centering
\setlength{\tabcolsep}{5pt}
\renewcommand{\arraystretch}{1.1}
\begin{tabular}{l|ccc|ccc}
\hline
      & \multicolumn{3}{c|}{MIA-A $\downarrow$ } & \multicolumn{3}{c}{MIA-C $\downarrow$ } \\
\cline{2-4} \cline{5-7}
Method & 01 & 05 & 10 & 01 & 05 & 10 \\
\hline
GA   & 0.72 & 0.76 & 0.80 & 0.83 & 0.86 & 0.88 \\
GD   & 0.73 & 0.78 & 0.80 & 0.83 & 0.86 & 0.88 \\
KL   & 0.70 & 0.76 & 0.80 & 0.82 & 0.86 & 0.88 \\
PO   & \underline{0.64} & \underline{0.69} & \underline{0.74} & \underline{0.78} & \underline{0.81} & \underline{0.84} \\
STaR & \textbf{0.51} & \textbf{0.53} & \textbf{0.53} & \textbf{0.62} & \textbf{0.65} & \textbf{0.68} \\
\hline
\end{tabular}
\caption{
Membership inference AUC (lower is better) for answer (MIA-A) and CoT (MIA-C) levels. 
}
\vspace{-0.2cm}
\label{tab:mia_metrics_auc}
\end{table}

\paragraph{Privacy Leakage under Membership Inference Attacks.}
Table~\ref{tab:mia_metrics_auc} further examines the privacy guarantees of each method by reporting AUC scores for membership inference attacks at both answer level (MIA-A) and CoT level (MIA-C), where lower values indicate better privacy protection. Consistently, STaR achieves the lowest AUC across all ratios and both attack types, substantially reducing privacy leakage risk compared to other methods. These results confirm that traditional unlearning approaches, even when effective in suppressing output similarity, are often vulnerable to adversarial attacks that exploit subtle statistical traces of forgotten data. In contrast, the dynamic, fine-grained suppression in STaR delivers strong privacy guarantees at multiple reasoning granularities.

\begin{table}[ht]
\centering
\setlength{\tabcolsep}{5pt}
\renewcommand{\arraystretch}{1.1}
\begin{tabular}{l|cc}
\hline
Method & Original Prompt $\uparrow$ & Paraphrased Prompt $\uparrow$ \\
\hline
GA   & 0.57 & 0.31 \\
GD   & 0.48 & 0.32 \\
KL   & 0.48 & 0.33 \\
PO   & \underline{0.68} & \underline{0.37} \\
STaR & \textbf{0.88} & \textbf{0.82} \\
\hline
\end{tabular}
\caption{
Answer-level forgetting efficacy (AFE) on the forget01 set under Original and paraphrased prompt queries. 
}
\vspace{-0.2cm}
\label{tab:paraphrase}
\end{table}

\paragraph{Robustness to Prompt Paraphrasing.}
To assess the robustness of unlearning methods against semantic variations of input queries, we evaluate forgetting efficacy on the forget set using both the original prompts and paraphrased versions generated by manually or automatically rewording the queries. As shown in Table~\ref{tab:paraphrase}, most baseline methods experience substantial drops in forgetting performance when presented with paraphrased prompts, highlighting their vulnerability to prompt rephrasing attacks. In contrast, STaR consistently maintains high forgetting efficacy, demonstrating strong resilience to both standard and semantically modified queries. This result underscores the practical reliability of our approach for privacy protection in real-world scenarios where user queries may vary in formulation.

\begin{table}[ht]
\centering
\setlength{\tabcolsep}{8pt}
\renewcommand{\arraystretch}{1.1}
\begin{tabular}{l|c}
\hline
Method &  Running Time (h) $\downarrow$  \\
\hline
GA      & 8.5 \\
GD      & 8.5 \\
KL      & \underline{6.0} \\
PO      & 9.0 \\
STaR    & \textbf{0.5} \\
\hline
\end{tabular}
\caption{
End-to-end running time (in hours) for each unlearning method on the R-TOFU benchmark.
}
\vspace{-0.2cm}
\label{tab:runtime}
\end{table}

\paragraph{Computational Efficiency.}
Table~\ref{tab:runtime} compares the total end-to-end running time of all methods on the R-TOFU benchmark under the forget10 protocol.All running time measurements are reported on the same  NVIDIA H800 GPU setup to ensure fairness. Retraining-based baselines (GA, GD, KL, PO) require multiple epochs of fine-tuning on large-scale models, resulting in high computational overhead. In contrast, our inference-time STaR framework completes the unlearning process in a fraction of the time, delivering over an order-of-magnitude speedup while maintaining superior forgetting efficacy and privacy protection. This substantial efficiency advantage highlights the practical deployability of STaR in real-world scenarios requiring prompt and reliable data deletion.

\begin{table}[!ht]
\centering
\resizebox{1.0\linewidth}{!}
  {
\begin{tabular}{l|cccc|cc}
\hline
Method      & MU$\uparrow$& AFE$\uparrow$& CFE$\uparrow$& MCS$\uparrow$& MIA-A$\downarrow$ & MIA-C$\downarrow$\\
\hline
w/o secure prompt & \textbf{0.93} & 0.82 & 0.66 & 0.65 & 0.56 & 0.6\\
w/o phrase   & 0.91 & 0.67 & 0.54 & 0.61 & 0.67  & 0.71  \\
hard only    & 0.92 & \textbf{0.93} & 0.62 & 0.69 & 0.70  & 0.68  \\
soft only    & 0.92 & 0.62 & 0.58 & 0.60 & 0.51  & \textbf{0.59}  \\
STaR (Full)  & \textbf{0.93} & \underline{0.88} & \textbf{0.68} & \textbf{0.70} & \textbf{0.51}  & \underline{0.62}  \\
\hline
\end{tabular}
}
\caption{
Ablation study of STaR on the forget01 split.
}
\vspace{-0.2cm}
\label{tab:ablation}
\end{table}

\paragraph{Ablation Study.}
Table~\ref{tab:ablation} reports ablation results for each core component of STaR. Removing semantic phrase expansion (\textit{w/o phrase}) significantly impairs both answer- and chain-level forgetting, and markedly increases membership inference risk, underscoring the need for high-order semantic coverage. Relying solely on hard suppression (\textit{hard only}) yields strong answer-level forgetting but fails to prevent privacy leakage or ensure multi-step consistency, while using only soft suppression (\textit{soft only}) weakens exact match blocking and results in inferior forgetting despite modest privacy gains. Notably, ablating the secure prompt module (\textit{w/o secure prompt}) leads to measurable declines in both chain-level forgetting (CFE) and decoding robustness (MCS), confirming that even soft, input-level constraints contribute to comprehensive privacy protection. Only the complete STaR pipeline, which integrates all modules, achieves uniformly strong performance across all metrics, highlighting the necessity of a holistic, multi-layered intervention strategy for robust and secure unlearning in LRMs.

\paragraph{Qualitative Analysis.}
Beyond quantitative improvements, qualitative results (see \textit{Appendix C}) highlights STaR's unique ability to suppress sensitive content across a broad spectrum of reasoning trajectories and adversarial decoding scenarios. For example, even when input prompts are strategically altered or reasoning steps are truncated , STaR consistently prevents the resurfacing of forgotten information, whereas baseline approaches often expose sensitive details via omitted or compressed reasoning. Representative case studies further illustrate that stepwise, decoding-agnostic intervention is essential for practical, end-to-end privacy protection in large reasoning models.


\section{Conclusion}
We present STaR, a novel inference-time unlearning framework for large reasoning models that achieves robust, decoding-agnostic suppression of sensitive information via semantic content identification, adaptive token-level suppression, and stepwise trajectory regulation. Extensive evaluation on the R-TOFU benchmark demonstrates that STaR consistently outperforms state-of-the-art baselines in forgetting efficacy and privacy protection at both the answer and chain-of-thought levels, while ablation studies confirm the indispensability of each module. Our results set a new benchmark for privacy-preserving unlearning in LRMs and provide a practical blueprint for compliant AI deployment.


\section*{Acknowledgments}

This work was supported by the National Nature Science Foundation of China (62322211), the "Pioneer" and "Leading Goose" R\&D Program of Zhejiang Province(2024C01023),Key Laboratory of Intelligent Processing Technology for Digital Music (Zhejiang Conservatory of Music), Ministry of Culture and Tourism (2023DMKLB004).This work was also supported by the National Nature Science Foundation of China (U25A20441: "Virtual--Physical Integrated Spatial Computing Theory and Methods for Complex Equipment Support").

\bibliography{aaai2026}

@String(PAMI = {IEEE Trans. Pattern Anal. Mach. Intell.})

@String(TIP  = {IEEE Trans. Image Process.})

@String(AAAI = {AAAI})

@String(PAMI  = {IEEE TPAMI})

@String(TIP   = {IEEE TIP})

@article{achiam2023gpt,
  title={Gpt-4 technical report},
  author={Achiam, Josh and Adler, Steven and Agarwal, Sandhini and Ahmad, Lama and Akkaya, Ilge and Aleman, Florencia Leoni and Almeida, Diogo and Altenschmidt, Janko and Altman, Sam and Anadkat, Shyamal and others},
  journal={arXiv preprint arXiv:2303.08774},
  year={2023}
}

@article{jaech2024openai,
  title={Openai o1 system card},
  author={Jaech, Aaron and Kalai, Adam and Lerer, Adam and Richardson, Adam and El-Kishky, Ahmed and Low, Aiden and Helyar, Alec and Madry, Aleksander and Beutel, Alex and Carney, Alex and others},
  journal={arXiv preprint arXiv:2412.16720},
  year={2024}
}

@article{guo2025deepseek,
  title={Deepseek-r1: Incentivizing reasoning capability in llms via reinforcement learning},
  author={Guo, Daya and Yang, Dejian and Zhang, Haowei and Song, Junxiao and Zhang, Ruoyu and Xu, Runxin and Zhu, Qihao and Ma, Shirong and Wang, Peiyi and Bi, Xiao and others},
  journal={arXiv preprint arXiv:2501.12948},
  year={2025}
}

@article{yoon2024rtofu,
  title={R-tofu: Unlearning in large reasoning models},
  author={Yoon, Sangyeon and Jeung, Wonje and No, Albert},
  journal={arXiv preprint arXiv:2505.15214},
  year={2025}
}

@article{wei2022chain,
  title={Chain-of-thought prompting elicits reasoning in large language models},
  author={Wei, Jason and Wang, Xuezhi and Schuurmans, Dale and Bosma, Maarten and Xia, Fei and Chi, Ed and Le, Quoc V and Zhou, Denny and others},
  journal={Advances in neural information processing systems},
  volume={35},
  pages={24824--24837},
  year={2022}
}

@inproceedings{carlini2021extracting,
  title={Extracting training data from large language models},
  author={Carlini, Nicholas and Tramer, Florian and Wallace, Eric and Jagielski, Matthew and Herbert-Voss, Ariel and Lee, Katherine and Roberts, Adam and Brown, Tom and Song, Dawn and Erlingsson, Ulfar and others},
  booktitle={30th USENIX security symposium (USENIX Security 21)},
  pages={2633--2650},
  year={2021}
}

@article{karamolegkou2023copyright,
  title={Copyright violations and large language models},
  author={Karamolegkou, Antonia and Li, Jiaang and Zhou, Li and S{\o}gaard, Anders},
  journal={arXiv preprint arXiv:2310.13771},
  year={2023}
}

@article{mireshghallah2023can,
  title={Can llms keep a secret? testing privacy implications of language models via contextual integrity theory},
  author={Mireshghallah, Niloofar and Kim, Hyunwoo and Zhou, Xuhui and Tsvetkov, Yulia and Sap, Maarten and Shokri, Reza and Choi, Yejin},
  journal={arXiv preprint arXiv:2310.17884},
  year={2023}
}

@article{maini2024tofu,
  title={Tofu: A task of fictitious unlearning for llms},
  author={Maini, Pratyush and Feng, Zhili and Schwarzschild, Avi and Lipton, Zachary C and Kolter, J Zico},
  journal={arXiv preprint arXiv:2401.06121},
  year={2024}
}

@article{rafailov2023direct,
  title={Direct preference optimization: Your language model is secretly a reward model},
  author={Rafailov, Rafael and Sharma, Archit and Mitchell, Eric and Manning, Christopher D and Ermon, Stefano and Finn, Chelsea},
  journal={Advances in neural information processing systems},
  volume={36},
  pages={53728--53741},
  year={2023}
}

@article{jiang2025safechain,
  title={Safechain: Safety of language models with long chain-of-thought reasoning capabilities},
  author={Jiang, Fengqing and Xu, Zhangchen and Li, Yuetai and Niu, Luyao and Xiang, Zhen and Li, Bo and Lin, Bill Yuchen and Poovendran, Radha},
  journal={arXiv preprint arXiv:2502.12025},
  year={2025}
}

@inproceedings{cao2015towards,
  title={Towards making systems forget with machine unlearning},
  author={Cao, Yinzhi and Yang, Junfeng},
  booktitle={2015 IEEE symposium on security and privacy},
  pages={463--480},
  year={2015},
  organization={IEEE}
}

@inproceedings{golatkar2020eternal,
  title={Eternal sunshine of the spotless net: Selective forgetting in deep networks},
  author={Golatkar, Aditya and Achille, Alessandro and Soatto, Stefano},
  booktitle={Proceedings of the IEEE/CVF conference on computer vision and pattern recognition},
  pages={9304--9312},
  year={2020}
}

@article{mekala2024alternate,
  title={Alternate preference optimization for unlearning factual knowledge in large language models},
  author={Mekala, Anmol and Dorna, Vineeth and Dubey, Shreya and Lalwani, Abhishek and Koleczek, David and Rungta, Mukund and Hasan, Sadid and Lobo, Elita},
  journal={arXiv preprint arXiv:2409.13474},
  year={2024}
}

@article{bhaila2024soft,
  title={Soft prompting for unlearning in large language models},
  author={Bhaila, Karuna and Van, Minh-Hao and Wu, Xintao},
  journal={arXiv preprint arXiv:2406.12038},
  year={2024}
}

@article{liu2024large,
  title={Large language model unlearning via embedding-corrupted prompts},
  author={Liu, Chris and Wang, Yaxuan and Flanigan, Jeffrey and Liu, Yang},
  journal={Advances in Neural Information Processing Systems},
  volume={37},
  pages={118198--118266},
  year={2024}
}

@article{kuo2025proactive,
  title={Proactive privacy amnesia for large language models: Safeguarding PII with negligible impact on model utility},
  author={Kuo, Martin and Zhang, Jingyang and Zhang, Jianyi and Tang, Minxue and DiValentin, Louis and Ding, Aolin and Sun, Jingwei and Chen, William and Hass, Amin and Chen, Tianlong and others},
  journal={arXiv preprint arXiv:2502.17591},
  year={2025}
}

@article{chen2023unlearn,
  title={Unlearn what you want to forget: Efficient unlearning for llms},
  author={Chen, Jiaao and Yang, Diyi},
  journal={arXiv preprint arXiv:2310.20150},
  year={2023}
}

@article{jia2024soul,
  title={Soul: Unlocking the power of second-order optimization for llm unlearning},
  author={Jia, Jinghan and Zhang, Yihua and Zhang, Yimeng and Liu, Jiancheng and Runwal, Bharat and Diffenderfer, James and Kailkhura, Bhavya and Liu, Sijia},
  journal={arXiv preprint arXiv:2404.18239},
  year={2024}
}

@article{yuan2024closer,
  title={A closer look at machine unlearning for large language models},
  author={Yuan, Xiaojian and Pang, Tianyu and Du, Chao and Chen, Kejiang and Zhang, Weiming and Lin, Min},
  journal={arXiv preprint arXiv:2410.08109},
  year={2024}
}

@article{zhang2024negative,
  title={Negative preference optimization: From catastrophic collapse to effective unlearning},
  author={Zhang, Ruiqi and Lin, Licong and Bai, Yu and Mei, Song},
  journal={arXiv preprint arXiv:2404.05868},
  year={2024}
}

@article{team2023gemini,
  title={Gemini: a family of highly capable multimodal models},
  author={Team, Gemini and Anil, Rohan and Borgeaud, Sebastian and Alayrac, Jean-Baptiste and Yu, Jiahui and Soricut, Radu and Schalkwyk, Johan and Dai, Andrew M and Hauth, Anja and Millican, Katie and others},
  journal={arXiv preprint arXiv:2312.11805},
  year={2023}
}

@article{touvron2023llama,
  title={Llama 2: Open foundation and fine-tuned chat models},
  author={Touvron, Hugo and Martin, Louis and Stone, Kevin and Albert, Peter and Almahairi, Amjad and Babaei, Yasmine and Bashlykov, Nikolay and Batra, Soumya and Bhargava, Prajjwal and Bhosale, Shruti and others},
  journal={arXiv preprint arXiv:2307.09288},
  year={2023}
}

@article{taylor2022galactica,
  title={Galactica: A large language model for science},
  author={Taylor, Ross and Kardas, Marcin and Cucurull, Guillem and Scialom, Thomas and Hartshorn, Anthony and Saravia, Elvis and Poulton, Andrew and Kerkez, Viktor and Stojnic, Robert},
  journal={arXiv preprint arXiv:2211.09085},
  year={2022}
}

@article{bao2024harnessing,
  title={Harnessing business and media insights with large language models},
  author={Bao, Yujia and Shah, Ankit Parag and Narang, Neeru and Rivers, Jonathan and Maksey, Rajeev and Guan, Lan and Barrere, Louise N and Evenson, Shelley and Basole, Rahul and Miao, Connie and others},
  journal={arXiv preprint arXiv:2406.06559},
  year={2024}
}

@inproceedings{zhang2024generate,
  title={To generate or not? safety-driven unlearned diffusion models are still easy to generate unsafe images... for now},
  author={Zhang, Yimeng and Jia, Jinghan and Chen, Xin and Chen, Aochuan and Zhang, Yihua and Liu, Jiancheng and Ding, Ke and Liu, Sijia},
  booktitle={European Conference on Computer Vision},
  pages={385--403},
  year={2024},
  organization={Springer}
}

@inproceedings{chu2024protect,
  title={How to protect copyright data in optimization of large language models?},
  author={Chu, Timothy and Song, Zhao and Yang, Chiwun},
  booktitle={Proceedings of the AAAI Conference on Artificial Intelligence},
  
  pages={17871--17879},
  year={2024},
  volume={38},
 
}

@article{staufer2025should,
  title={What Should LLMs Forget? Quantifying Personal Data in LLMs for Right-to-Be-Forgotten Requests},
  author={Staufer, Dimitri},
  journal={arXiv preprint arXiv:2507.11128},
  year={2025}
}

@article{xiong2025landscape,
  title={The Landscape of Memorization in LLMs: Mechanisms, Measurement, and Mitigation},
  author={Xiong, Alexander and Zhao, Xuandong and Pappu, Aneesh and Song, Dawn},
  journal={arXiv preprint arXiv:2507.05578},
  year={2025}
}

@article{scholten2025model,
  title={Model Collapse Is Not a Bug but a Feature in Machine Unlearning for LLMs},
  author={Scholten, Yan and Xhonneux, Sophie and G{\"u}nnemann, Stephan and Schwinn, Leo},
  journal={arXiv preprint arXiv:2507.04219},
  year={2025}
}

@article{zhao2025improving,
  title={Improving llm safety alignment with dual-objective optimization},
  author={Zhao, Xuandong and Cai, Will and Shi, Tianneng and Huang, David and Lin, Licong and Mei, Song and Song, Dawn},
  journal={arXiv preprint arXiv:2503.03710},
  year={2025}
}

@article{feng2025existing,
  title={Existing Large Language Model Unlearning Evaluations Are Inconclusive},
  author={Feng, Zhili and Xu, Yixuan Even and Robey, Alexander and Kirk, Robert and Davies, Xander and Gal, Yarin and Schwarzschild, Avi and Kolter, J Zico},
  journal={arXiv preprint arXiv:2506.00688},
  year={2025}
}

@article{reisizadeh2025blur,
  title={BLUR: A Bi-Level Optimization Approach for LLM Unlearning},
  author={Reisizadeh, Hadi and Jia, Jinghan and Bu, Zhiqi and Vinzamuri, Bhanukiran and Ramakrishna, Anil and Chang, Kai-Wei and Cevher, Volkan and Liu, Sijia and Hong, Mingyi},
  journal={arXiv preprint arXiv:2506.08164},
  year={2025}
}

@article{wan2025not,
  title={Not Every Token Needs Forgetting: Selective Unlearning to Limit Change in Utility in Large Language Model Unlearning},
  author={Wan, Yixin and Ramakrishna, Anil and Chang, Kai-Wei and Cevher, Volkan and Gupta, Rahul},
  journal={arXiv preprint arXiv:2506.00876},
  year={2025}
}

@article{chen2025does,
  title={Does Machine Unlearning Truly Remove Model Knowledge? A Framework for Auditing Unlearning in LLMs},
  author={Chen, Haokun and Zhang, Yueqi and Bi, Yuan and Zhang, Yao and Liu, Tong and Bi, Jinhe and Lan, Jian and Gu, Jindong and Grosser, Claudia and Krompass, Denis and others},
  journal={arXiv preprint arXiv:2505.23270},
  year={2025}
}

@article{to2025harry,
  title={Harry potter is still here! probing knowledge leakage in targeted unlearned large language models via automated adversarial prompting},
  author={To, Bang Trinh Tran and Le, Thai},
  journal={arXiv preprint arXiv:2505.17160},
  year={2025}
}

@article{xu2025obliviate,
  title={OBLIVIATE: Robust and Practical Machine Unlearning for Large Language Models},
  author={Xu, Xiaoyu and Du, Minxin and Ye, Qingqing and Hu, Haibo},
  journal={arXiv preprint arXiv:2505.04416},
  year={2025}
}

@article{sun2024learning, title={Learning and unlearning of fabricated knowledge in language models}, author={Sun, Chen and Miller, Nolan Andrew and Zhmoginov, Andrey and Vladymyrov, Max and Sandler, Mark}, journal={arXiv preprint arXiv:2410.21750}, year={2024} }

@article{jia2024wagle, title={Wagle: Strategic weight attribution for effective and modular unlearning in large language models}, author={Jia, Jinghan and Liu, Jiancheng and Zhang, Yihua and Ram, Parikshit and Baracaldo, Nathalie and Liu, Sijia}, journal={Advances in Neural Information Processing Systems}, volume={37}, pages={55620--55646}, year={2024} }

@article{sinha2024unstar, title={Unstar: Unlearning with self-taught anti-sample reasoning for llms}, author={Sinha, Yash and Mandal, Murari and Kankanhalli, Mohan}, journal={arXiv preprint arXiv:2410.17050}, year={2024} }

@article{wang2024machine, title={When machine unlearning meets retrieval-augmented generation (rag): Keep secret or forget knowledge?}, author={Wang, Shang and Zhu, Tianqing and Ye, Dayong and Zhou, Wanlei}, journal={arXiv preprint arXiv:2410.15267}, year={2024} }

@article{li2025dubbing,
  author    = {Liang Li and Gaoxiang Cong and Yuankai Qi and Zheng-Jun Zha and Qi Wu and Michael Sheng and Qingming Huang and Ming-Hsuan Yang},
  title     = {Dubbing Movies via Hierarchical Phoneme Modeling and Acoustic Diffusion Denoising},
  journal   = PAMI,
  year      = {2025}
}

@article{zhang2024inductive,
  author    = {Beichen Zhang and Liang Li and Shuhui Wang and Shaofei Cai and Zheng-Jun Zha and Qi Tian and Qingming Huang},
  title     = {Inductive State-Relabeling Adversarial Active Learning with Heuristic Clique Rescaling},
  journal   = PAMI,
  year      = {2024}
}

@article{tu2024smart,
  author    = {Yunbin Tu and Liang Li and Li Su and Zheng-Jun Zha and Qingming Huang},
  title     = {SMART: Syntax-Calibrated Multi-Aspect Relation Transformer for Change Captioning},
  journal   = PAMI,
  year      = {2024}
}

@article{liu2023entity,
  author    = {Xuejing Liu and Liang Li and Shuhui Wang and Zheng-Jun Zha and Zechao Li and Qi Tian and Qingming Huang},
  title     = {Entity-Enhanced Adaptive Reconstruction Network for Weakly Supervised Referring Expression Grounding},
  journal   = PAMI,
  volume    = {45},
  number    = {3},
  pages     = {3003--3018},
  year      = {2023}
}

@article{li2022lstt,
  author    = {Liang Li and Xingyu Gao and Jincan Deng and Yunbin Tu and Zheng-Jun Zha and Qingming Huang},
  title     = {Long Short-Term Relation Transformer with Global Gating for Video Captioning},
  journal   = TIP,
  year      = {2022}
}

\end{document}